\documentclass[letterpaper, 10 pt, journal]{IEEEtran}  

\usepackage{amsmath}
\usepackage{graphicx}
\usepackage{cite}
\usepackage{xcolor}

\title{\LARGE \bf
A Compact Robotic Finger with 2-DoF MCP Joint Embedding DoF-Selective Passive Continuously Variable Transmission for Wide Force–Speed Operating Range}

\author{JaeHyung Jang and Jee-Hwan Ryu%
\thanks{JaeHyung Jang and Jee-Hwan Ryu are with the Department of Civil and Environmental Engineering, Korea Advanced Institute of Science and Technology, Daejeon 34141, South Korea. (e-mail: jhjang.kd@kaist.ac.kr; jhryu@kaist.ac.kr)}%
}

\begin{document}

\maketitle
\thispagestyle{empty}
\pagestyle{empty}

\begin{abstract}
This letter presents a compact two-degree-of-freedom (DoF) robotic finger with a flexion-selective passive continuously variable transmission (CVT) to achieve a wide force–speed operating range. Inspired by the functional differentiation of the human metacarpophalangeal (MCP) joint, the proposed mechanism realizes DoF-specific transmission differentiation by selectively assigning passive CVT to the flexion–extension DoF while preserving direct transmission for abduction–adduction. For a wide force–speed operating range, a force-responsive passive CVT is embedded in the flexion pathway, while direct transmission is preserved for the abduction–adduction pathway. To selectively realize transmission adaptation within a multi-DoF MCP mechanism, an output-side passive CVT employing a moving-pulley-inspired wire-routing structure is introduced. The resultant force generated by the wire tensions acting on the pulley that passively increases the flexion moment arm and transmission ratio according to the applied load without additional actuators, sensors, or control. Experimental results demonstrate a maximum output-force amplification of 4.19-fold and a mean amplification of 3.6-fold across the tested flexion angles ranging from 15° to 75° through moment-arm adaptation, thereby substantially expanding the achievable force–-speed operating range. Furthermore, dexterous ball-rolling experiments verify that passive transmission adaptation can be achieved while preserving abduction–adduction functionality. These results demonstrate a scalable transmission design strategy for compact multi-DoF robotic hands.

\end{abstract}

\section{INTRODUCTION}

The human hand achieves remarkable dexterity by simultaneously realizing delicate manipulation, powerful grasping, and rapid finger motion within a compact musculoskeletal structure~\cite{bicchi2000hands}. This capability enables versatile object manipulation and precise force regulation in unstructured environments and has long served as a benchmark for robotic hand development~\cite{zhou2024dexterous}. However, achieving these performance requirements within compact robotic hands remains challenging. In in-finger architectures, limited installation volume fundamentally constrains actuator output, making it difficult to simultaneously achieve high grasping force and rapid joint motion. These mechanically conflicting requirements create an inherent force-speed tradeoff and represent a major limitation in compact multi-DoF robotic hands~\cite{kim2021integrated}.

Tendon-driven actuation and variable transmission (VT) have been extensively investigated to address this challenge. Tendon-driven architectures relocate actuators to the forearm and transmit force through wires, enabling the use of actuators with substantially higher output than can be embedded within the finger~\cite{xu2016design,kim2019fluid,zhang2025biomimetic}. However, wire paths crossing the wrist joint introduce motion coupling between wrist posture and finger motion, while accumulated friction along extended routing paths degrades force transmission efficiency and control precision. These effects can complicate accurate force and motion transmission in compact multi-DoF robotic hands.
\begin{table*}[!t]
\centering
\caption{Comparison of variable-transmission mechanisms applied to robotic fingers.}
\label{tab:vt_comparison}

\renewcommand{\arraystretch}{0.95}
\setlength{\tabcolsep}{2.0pt}
\fontsize{7}{8}\selectfont

\begin{tabular*}{\textwidth}{@{\extracolsep{\fill}}lcccccc@{}}
\hline

\textbf{Work} &
\textbf{\begin{tabular}[c]{@{}c@{}}
Reported Transmission\\
Variation
\end{tabular}} &
\textbf{\begin{tabular}[c]{@{}c@{}}
Transmission\\
Type
\end{tabular}} &
\textbf{Adaptation} &
\textbf{\begin{tabular}[c]{@{}c@{}}
Additional VT\\
Actuator
\end{tabular}} &
\textbf{\begin{tabular}[c]{@{}c@{}}
VT\\
Integration
\end{tabular}} &
\textbf{\begin{tabular}[c]{@{}c@{}}
VT-Adapted DoFs /\\
Motion DoFs$^{**}$
\end{tabular}} \\

\hline

Jeong et al. \cite{jeong2017designing} &
6.9$\times$ (speed) / 11.6$\times$ (torque) &
Discrete &
Active &
1 &
Actuator-side &
1 / 1 \\

Spanjer et al. \cite{spanjer2012underactuated} &
N/R$^{*}$ &
Continuous &
Active &
1 &
Actuator-side &
1 / 1 \\

Chung et al. \cite{chung2024robotic} &
2.13$\times$ &
Continuous &
Active &
1 &
MCP-joint integrated &
1 / 1 \\

\hline

Phlernjai et al. \cite{phlernjai2016passively} &
9.1$\times$ &
Discrete &
Passive &
0 &
Actuator-side &
1 / 1 \\

Matsushita et al. \cite{matsushita2009development} &
2--3$\times$ &
Continuous &
Passive &
0 &
Actuator-side &
1 / 1 \\

O'Brien et al. \cite{o2018elastomeric} &
2.88$\times$ &
Continuous &
Passive &
0 &
Actuator-side &
1 / 1 \\

\textbf{STAND-MCP (ours)} &
\textbf{4.19$\times$ (max) / 3.63$\times$ (mean)} &
\textbf{Continuous} &
\textbf{Passive} &
\textbf{0} &
\textbf{MCP-joint integrated} &
\textbf{1 / 2} \\

\hline
\end{tabular*}

\vspace{0.3mm}

\begin{minipage}{\textwidth}
\tiny
$^{*}$N/R indicates that a quantitative transmission-variation ratio was not explicitly reported in the corresponding study; reported transmission variations follow the transmission-performance metrics defined in the respective studies.\\
$^{**}$VT-Adapted DoFs / Motion DoFs represent the number of functional motion DoFs incorporating VT relative to the total functional motion DoFs per finger, excluding internal transmission-reconfiguration DoFs.
\end{minipage}

\end{table*}

Variable transmission mitigates the force-speed tradeoff by adapting the transmission ratio according to operating conditions. Active VT mechanisms achieve transmission adaptation through dedicated actuators, sensors, and control, providing versatile transmission characteristics~\cite{jeong2017design}.
Such approaches have been applied to robotic fingers, including a compact continuously variable active transmission based on a four-bar linkage~\cite{chung2024robotic} and a variable-radius pulley for an underactuated finger~\cite{spanjer2012underactuated}. However, active transmission adaptation requires additional actuation and reconfiguration components, increasing the mechanical and packaging requirements of the finger system.

Passive VT mechanisms reduce the need for additional transmission actuation by adapting the transmission ratio according to the applied load or mechanical state, offering advantages in compactness and fast responsiveness~\cite{belter2014passively}. Passive transmission adaptation has also been demonstrated in robotic fingers and hands through load-sensitive cable-driven transmissions, drum-type CVTs, and
elastomeric passive transmissions~\cite{matsushita2009development,o2018elastomeric,phlernjai2016passively}. 
Nevertheless, when variable-transmission functionality is extended to multi-DoF robotic fingers, each DoF requiring transmission adaptation generally requires its own transmission-related mechanical elements and corresponding installation space. Consequently, although active and passive VT architectures differ in how transmission adaptation is realized, both can incur increasing mechanical and packaging burdens when VT functionality is extended across multiple DoFs in compact robotic fingers.

In the human hand, the metacarpophalangeal (MCP) joint possesses a parallel two-DoF architecture consisting of flexion--extension (FE) and abduction--adduction (AA)~\cite{charles1970intrinsic}. These 2-DoFs are functionally differentiated and therefore impose different performance requirements. This functional differentiation provides an opportunity to design robotic hands according to the role of each DoF rather than assigning identical transmission characteristics across all DoFs.

Based on this observation, we present the DoF-Specific TrANsmission Differentiation-based passive CVT-embedded MCP (STAND-MCP) mechanism, in which passive variable-transmission functionality is embedded directly within the MCP joint structure. The joint-integrated architecture selectively provides force-responsive transmission adaptation to the FE pathway while retaining direct transmission for AA, thereby avoiding an additional variable-transmission element in the AA pathway.

Table~\ref{tab:vt_comparison} compares representative variable-transmission mechanisms applied to robotic fingers in terms of transmission characteristics, adaptation requirements, integration strategy, and the functional motion DoFs to which variable-transmission is applied. Among the representative finger-level variable-transmission mechanisms considered in the comparison, variable-transmission is applied to the single functional motion DoF of each finger, whereas the STAND-MCP preserves two functional motion DoFs while selectively applying variable-transmission only to flexion. Furthermore, the passive CVT is integrated directly within the MCP joint and requires no additional actuator for transmission adaptation. This combination of DoF-selective transmission adaptation and joint-integrated implementation reduces the transmission-related mechanical and packaging burden associated with introducing VT into a multi-DoF finger architecture.

To realize this architecture within the limited MCP joint volume, an output-side passive CVT based on a moving-pulley mechanism is introduced. The moving pulley is incorporated directly into the MCP transmission structure and changes its position according to wire tension, thereby continuously varying the FE moment arm and transmission ratio without additional sensing or active mode control.

This architecture addresses the mechanical and packaging burden associated with extending variable-transmission functionality to multi-DoF robotic fingers through two complementary features: selective transmission adaptation and joint-integrated implementation. Passive CVT functionality is applied to flexion, where a wide force-speed operating range is required, while abduction--adduction retains direct transmission, and the transmission-reconfiguration components are integrated directly within the MCP joint through the moving-pulley mechanism. The resulting STAND-MCP mechanism is theoretically and experimentally validated, demonstrating force-responsive adaptation of the FE transmission ratio, expansion of the flexion force-speed operating range, and coupled FE--AA motion within a compact 2-DoF MCP mechanism.

The remainder of this paper is organized as follows. Section~\ref{sec:section2} introduces the design concept and working principle of the STAND-MCP. Section~\ref{sec:section3} presents the theoretical modeling of the underactuated mechanism, including bilateral wire kinematics, the flexion-specific passive CVT, passive extension, and condition-dependent motion generation. Section~\ref{sec:section4} describes the mechanical design and parameter selection of the proposed mechanism. Section~\ref{sec:section5} experimentally validates the transmission kinematics and force-responsive transmission behavior of the proposed passive CVT. Section~\ref{sec:section6} demonstrates the applicability of the mechanism to dexterous robotic manipulation. Finally, Section~\ref{sec:section7} concludes the paper and discusses future research directions.

\section{Concept of the STAND-MCP}
\label{sec:section2}

\begin{figure*}[t]
    \centering
    \includegraphics[width=0.95\textwidth]{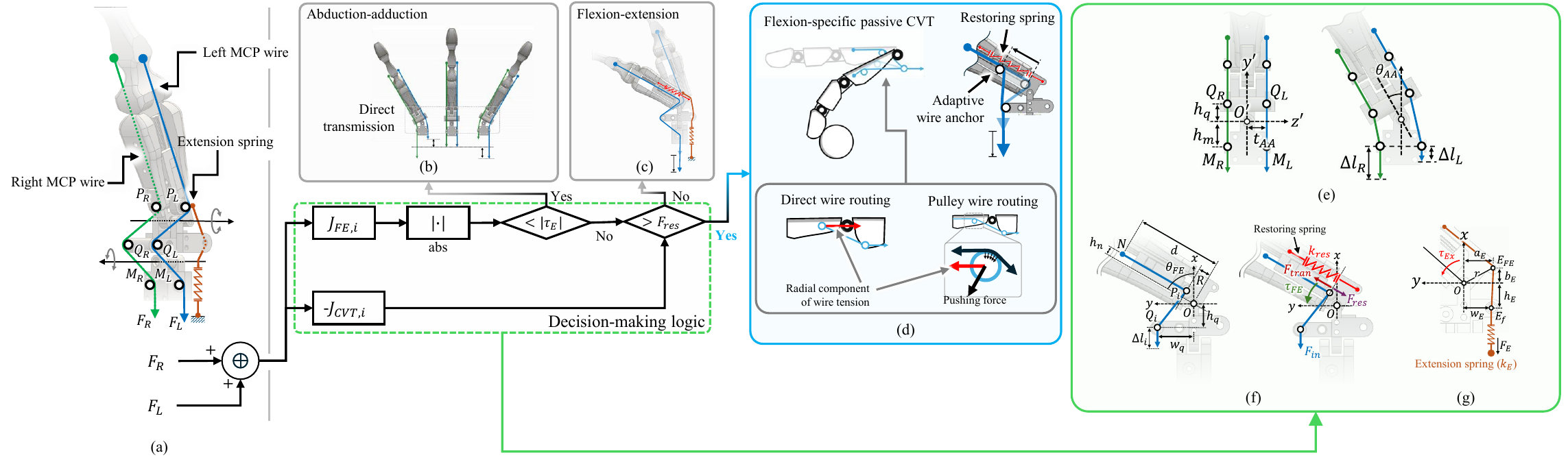}
    \caption{Concept and decision-making logic of the STAND-MCP. (a) Overview of the STAND-MCP. (b) Direct transmission for abduction–adduction. (c) Flexion–extension motion with a flexion-specific passive CVT. (d) Force-responsive passive CVT employing a moving-pulley-inspired wire-routing structure. (e) Bilateral wire-routing geometry for abduction–adduction motion. (f) Wire-routing geometry of the flexion and flexion-specific passive CVT. (g) Passive extension mechanism geometry.}
    \label{fig:concept}
\end{figure*}

\subsection{Concept and Working Principle}
The human finger consists of three joints: the metacarpophalangeal (MCP), proximal interphalangeal, and distal interphalangeal, providing two, one, and one degrees of freedom (DoFs), respectively. Among these, the MCP joint is the finger joint that possesses a parallel two-DoF architecture, enabling both flexion--extension (FE) and abduction--adduction (AA). These 2-DoFs perform distinct mechanical roles and therefore impose different performance requirements. Flexion--extension requires a wide force-speed operating range to achieve rapid finger motion before contact and strong grasping after contact. In contrast, abduction--adduction governs inter-finger spacing and object alignment, making compactness and positional controllability its primary design requirements.

The STAND-MCP mechanism is designed based on this principle of functional differentiation and realizes DoF-specific transmission differentiation (Fig.~\ref{fig:concept}(a)). A force-responsive passive CVT is selectively applied to the flexion pathway, which demands a wide force-speed operating range, whereas direct transmission is retained for the abduction-adduction pathway (Fig.~\ref{fig:concept}(b)), where compactness and positional controllability are prioritized. Through this design, the STAND-MCP achieves selective variable transmission within a compact multi-DoF MCP architecture.

The flexion-specific passive CVT embedded in the STAND-MCP is designed to passively shift the adaptive wire anchor according to wire tension, thereby reconfiguring the flexion moment arm and mechanically modulating the transmission ratio (Fig.~\ref{fig:concept}(c)). Achieving passive reconfiguration of the moment arm by wire tension requires the radial component of the wire tension to act away from the joint center. However, direct wire routing to the adaptive anchor, as illustrated in Fig.~\ref{fig:concept}(d), causes the radial component of the wire tension to act toward the joint center. This effect results in a decrease in the flexion moment arm.

To reverse this force direction, a moving-pulley structure in which the wire is routed around a pulley is adopted. In this structure, the resultant force generated by the wire tensions acting on the pulley is converted into a pushing force that drives the adaptive anchor away from the joint center. Because this pushing force grows together with the flexion wire tension, an increase in flexion resistance caused by external contact induces passive displacement of the adaptive anchor and expansion of the flexion moment arm. As a result, the passive CVT element is mechanically assigned to the flexion transmission pathway, whereas the abduction--adduction pathway retains direct transmission, thereby realizing DoF-specific transmission differentiation within a single MCP joint.


The STAND-MCP is an underactuated system in which three degrees of freedom, two parallel DoFs for FE and AA motion and one DoF for the flexion-specific passive CVT, are controlled by two actuators. Owing to this underactuation, the overall behavior of the system is determined by the combination of the two actuator inputs applied to the left and right wires, making it a mechanically programmed structure in which the decision-making logic is inherently embedded in the mechanical architecture itself \cite{jang2026morph}. The conditions of this logic are determined by the kinematics and kinetics of the system together with the stiffness of the extension spring, which enables passive extension, and the restoring spring, which biases the adaptive anchor toward its initial position. In this section, we analyze the kinematics and kinetics of flexion including the flexion-specific passive CVT, passive extension, and AA to derive these conditions, and define the resulting motions according to the derived conditions.

\section{Design and Modeling}
\label{sec:section3}

\subsection{Bilateral Wire Kinematics and Flexion-Specific Passive CVT}
\label{sec:bilateral_kinematics}

The STAND-MCP is actuated by bilateral wires whose routing geometry
varies with both flexion--extension (FE) and abduction--adduction (AA)
motion. In addition, the passive continuously variable transmission
(CVT) changes the effective flexion moment arm $R$. Accordingly, the
generalized coordinate vector is defined as
\begin{equation}
\mathbf{x}
=
\begin{bmatrix}
\theta_{FE} &
\theta_{AA} &
R
\end{bmatrix}^{T}.
\label{eq:generalized_coordinate}
\end{equation}
The left and right wire paths are indexed by $i\in\{L,R\}$. As shown
in Fig.~\ref{fig:concept}(e) and (f), $\mathbf{M}_i$ is fixed to the base,
$\mathbf{Q}_i$ varies with $\theta_{AA}$, and $\mathbf{P}_i$ depends
on $\theta_{FE}$, $\theta_{AA}$, and $R$:
\begin{equation}
\mathbf{M}_i=\mathrm{const.},\;
\mathbf{Q}_i=\mathbf{Q}_i(\theta_{AA}),\;
\mathbf{P}_i=\mathbf{P}_i(\theta_{FE},\theta_{AA},R).
\label{eq:routing_points}
\end{equation}
Since the wire segment between $\mathbf{N}_i$ and $\mathbf{P}_i$ is
$\lvert d-R \rvert$, the total wire length is expressed as
\begin{equation}
\begin{aligned}
L_i(\theta_{FE},\theta_{AA},R)
={}&
(\lvert d-R \rvert)
+
\|\mathbf{P}_i-\mathbf{Q}_i\|
\\
&+
\|\mathbf{Q}_i-\mathbf{M}_i\|
+
L_{i,\mathrm{const}}.
\end{aligned}
\label{eq:total_wire_length}
\end{equation}
where $L_{i,\mathrm{const}}$ represents the wire segments whose lengths
are unaffected by the MCP motion. Compared with the previous planar
formulation, (\ref{eq:total_wire_length}) explicitly includes the
AA-dependent variation of the $\mathbf{Q}_i$--$\mathbf{M}_i$ segment.

The wire displacement relative to the neutral configuration is
\begin{equation}
\Delta l_i
=
L_i(0,0,R_0)
-
L_i(\theta_{FE},\theta_{AA},R).
\label{eq:wire_displacement}
\end{equation}
The wire-length Jacobian is defined as
\begin{equation}
\mathbf{J}_i
=
\frac{\partial L_i}{\partial\mathbf{x}}
=
\begin{bmatrix}
J_{FE,i} &
J_{AA,i} &
J_{CVT,i}
\end{bmatrix},
\label{eq:wire_jacobian}
\end{equation}
where
\begin{equation}
J_{FE,i}=\frac{\partial L_i}{\partial\theta_{FE}},\quad
J_{AA,i}=\frac{\partial L_i}{\partial\theta_{AA}},\quad
J_{CVT,i}=\frac{\partial L_i}{\partial R}.
\label{eq:jacobian_components}
\end{equation}
In particular, because AA motion changes both $\mathbf{P}_i$ and
$\mathbf{Q}_i$, the AA Jacobian contains contributions from both
variable wire segments:
\begin{equation}
J_{AA,i}
=
\frac{\partial\|\mathbf{P}_i-\mathbf{Q}_i\|}
{\partial\theta_{AA}}
+
\frac{\partial\|\mathbf{Q}_i-\mathbf{M}_i\|}
{\partial\theta_{AA}}.
\label{eq:AA_jacobian}
\end{equation}
Thus, the left and right AA Jacobians are generally not equal in
magnitude at the same nonzero AA angle. For the bilaterally
mirror-symmetric routing geometry,
\begin{equation}
J_{AA,L}(\theta_{FE},\theta_{AA},R)
=
-
J_{AA,R}(\theta_{FE},-\theta_{AA},R),
\label{eq:AA_symmetry}
\end{equation}
which reduces to $J_{AA,L}=-J_{AA,R}$ at $\theta_{AA}=0$. This relation indicates that the mechanical advantage of the left and right wire paths can become asymmetric at nonzero AA postures, while the symmetric transmission relationship is recovered at the neutral AA configuration.

Let $F_L$ and $F_R$ denote the left and right wire tensions,
respectively. The wire-generated transmission force acting on the
passive CVT coordinate is
\begin{equation}
F_{tran}
=
-
\left(
J_{CVT,L}F_L
+
J_{CVT,R}F_R
\right).
\label{eq:transmission_force}
\end{equation}
Here, the positive $R$ direction is defined as the direction of increasing 
flexion moment arm. Accordingly, $J_{CVT,i}<0$ over the operating range.

The slider restoring force is modeled as
\begin{equation}
F_{res}
=
F_{res,0}
+
k_{res}(R-R_0).
\label{eq:restoring_force}
\end{equation}
To account for Coulomb friction at the slider, the friction force is
modeled from the component of the wire tensions normal to the slider
motion. Defining
$\mathbf{u}_i=(\mathbf{Q}_i-\mathbf{P}_i)/
\|\mathbf{Q}_i-\mathbf{P}_i\|$ as the unit wire direction at
$\mathbf{P}_i$ and $\mathbf{n}_R$ as the unit vector normal to the
slider motion, the friction force is expressed as
\begin{equation}
F_{fric}^{\pm}
=
\mu^{\pm}
\sum_{i\in\{L,R\}}
F_i
\left|
\mathbf{u}_i^{T}\mathbf{n}_R
\right|,
\label{eq:friction_force}
\end{equation}
where $\mu^{+}$ and $\mu^{-}$ denote the effective Coulomb friction
coefficients during loading and unloading, respectively. The
quasi-static force balance is therefore
\begin{equation}
F_{tran}
=
\begin{cases}
F_{res}+F_{fric}^{+}, & \text{loading},\\
F_{res}-F_{fric}^{-}, & \text{unloading}.
\end{cases}
\label{eq:friction_force_balance}
\end{equation}
Combining (\ref{eq:restoring_force}) and
(\ref{eq:friction_force_balance}), the flexion moment arm is determined
by the quasi-static equilibrium as
\begin{equation}
R
=
R_0
+
\frac{1}{k_{res}}
\begin{cases}
F_{tran}-F_{fric}^{+}-F_{res,0}, & \text{loading},\\
F_{tran}+F_{fric}^{-}-F_{res,0}, & \text{unloading}.
\end{cases}
\label{eq:R_equilibrium}
\end{equation}
As the wire-generated transmission force exceeds the combined restoring and friction forces, the slider moves in the positive $R$ direction, increasing the flexion moment arm and thereby increasing the transmission ratio.
Because both $F_{tran}$ and $F_{fric}^{\pm}$ depend on the instantaneous wire-routing geometry and wire tension, (\ref{eq:R_equilibrium}) generally represents an implicit quasi-static equilibrium relation for $R$.

Although the instantaneous wire-routing geometry is affected by both
FE and AA posture, the mechanically reconfigurable passive CVT element
is assigned to the FE transmission pathway, whereas the AA pathway
retains direct transmission. Thus, the proposed mechanism realizes
DoF-specific transmission differentiation without assuming complete
kinematic decoupling between FE and AA.

To separately characterize the transmission performance of the
flexion-specific passive CVT, the FE transmission ratio is defined
under the neutral AA configuration ($\theta_{AA}=0$). This reference
configuration isolates the transmission variation associated with
the flexion moment arm $R$ from the posture-dependent wire-routing
variation introduced by AA motion. Accordingly,
\begin{equation}
TR_{FE}(\theta_{FE},R)
=
J_{FE,i}(\theta_{FE},0,R).
\label{eq:TR_FE_i}
\end{equation}
Here, $TR_{FE}$ represents the transmission characteristic of the
flexion-specific passive CVT and is used to quantify the transmission
variation produced by passive moment-arm adaptation. In contrast,
$J_{FE,i}$ in (\ref{eq:jacobian_components}) retains the general
posture dependence required to describe the complete bilateral
FE--AA wire kinematics.

\subsection{Kinematics and Kinetics of Passive Extension}
\label{sec:passive_extension}

Passive extension of the proposed mechanism is realized through an extension spring that generates a restoring torque against flexion motion (Fig.~\ref{fig:concept}(f)). This section derives the geometry-dependent extension torque model that characterizes the passive restoring behavior of the STAND-MCP.
A fixed spring anchor point $\mathbf{E}_f$ is defined in the global frame as:

\begin{equation}
\mathbf{E}_f
=
\begin{bmatrix}
-w_E \\
-h_E
\end{bmatrix}
\label{eq:spring_anchor}
\end{equation}
while the spring attachment point $\mathbf{E}_{FE}$, defined in the global frame, is expressed as:

\begin{equation}
\mathbf{E}_{FE}(\theta_{FE})
=
\mathbf{R}_z(\theta_{FE})
\begin{bmatrix}
a_E \\
b_E
\end{bmatrix},
\label{eq:spring_attachment_global}
\end{equation}
where $\mathbf{R}_z(\theta_{FE})$ denotes the planar rotation matrix. The spring direction vector $\mathbf{d}_E$ and instantaneous spring length $L_s(\theta_{FE})$ are then obtained as:

\begin{equation}
\mathbf{d}_E(\theta_{FE})
=
\mathbf{E}_f
-
\mathbf{E}_{FE}(\theta_{FE})
\label{eq:spring_direction}
\end{equation}

\begin{equation}
L_s(\theta_{FE})
=
\left\|
\mathbf{d}_E(\theta_{FE})
\right\|
\label{eq:spring_length}
\end{equation}
where the natural spring length is defined at the neutral configuration $(\theta_{FE}=0)$ as:

\begin{equation}
L_{s,0}
=
L_s(0)
\label{eq:natural_length}
\end{equation}
Assuming a linear spring with stiffness $k_E$ and preload force $F_{E,0}$, the spring force magnitude is expressed as:

\begin{equation}
F_E(\theta_{FE})
=
k_E
\left(
L_s(\theta_{FE})
-
L_{s,0}
\right)
+
F_{E,0}
\label{eq:spring_force}
\end{equation}
Because the spring force acts along the spring direction, the corresponding force vector becomes:

\begin{equation}
\mathbf{F}_E(\theta_{FE})
=
\frac{
F_E(\theta_{FE})
}{
L_s(\theta_{FE})
}
\mathbf{d}_E(\theta_{FE}).
\label{eq:spring_force_vector}
\end{equation}

The scalar extension torque about the FE axis is given by:
\begin{equation}
\begin{aligned}
\tau_{E}(\theta_{FE})
&=
\mathbf{E}_{FE}(\theta_{FE})
\times
\mathbf{F}_E(\theta_{FE})
\\
&=
\frac{
F_E(\theta_{FE})
}{
L_s(\theta_{FE})
}
\left(
w_E y_E
-
h_E x_E
\right)
\end{aligned}
\label{eq:extension_torque}
\end{equation}
Equation~\eqref{eq:extension_torque} indicates that the passive extension torque is governed by the spring routing geometry together with the spring stiffness and preload.

{
\subsection{Condition-Dependent Motion Generation}
\label{sec:condition_motion}

The STAND-MCP mechanism constitutes an underactuated system with two
wire inputs and three mechanically coupled internal coordinates:
flexion--extension (FE), abduction--adduction (AA), and passive
moment-arm reconfiguration through the force-responsive slider
(Fig.~\ref{fig:concept}). While the wire kinematics, passive CVT
behavior, and extension spring model were derived in the preceding
sections, the actual motion generated by the mechanism is determined
by the force balance established among these elements. Specifically,
the bilateral wire tensions selectively activate different mechanical
states according to the restoring torque of the extension spring and
the restoring force acting on the passive slider.

Using the posture-dependent bilateral Jacobians derived in
Section~\ref{sec:bilateral_kinematics}, the generalized outputs of the
proposed mechanism can be expressed as

\begin{equation}
{\setlength{\arraycolsep}{1pt}
\begin{bmatrix}
\tau_{FE}\\
F_{CVT}\\
\tau_{AA}
\end{bmatrix}
=
\begin{bmatrix}
J_{FE,L} & J_{FE,R}\\
-J_{CVT,L} & -J_{CVT,R}\\
J_{AA,L} & J_{AA,R}
\end{bmatrix}
\begin{bmatrix}
F_L\\
F_R
\end{bmatrix}
+
\begin{bmatrix}
-\tau_E\\
-F_{res}\\
0
\end{bmatrix}
}
\label{eq:generalized_output_matrix}
\end{equation}
where $\tau_E$ is the extension restoring torque and $F_{res}$ is the
restoring force acting on the passive slider.
For clarity, the following mode-transition conditions are formulated
under an idealized quasi-static assumption neglecting slider friction.

The actuation mode is determined by the following force conditions:

\begin{gather}
\left|
J_{FE,L}F_L
+
J_{FE,R}F_R
\right|
<
|\tau_E|
\label{eq:condition_AA}
\\
\left|
J_{FE,L}F_L
+
J_{FE,R}F_R
\right|
>
|\tau_E|
\land
F_{tran}
<
F_{res}
\label{eq:condition_FE}
\\
\left|
J_{FE,L}F_L
+
J_{FE,R}F_R
\right|
>
|\tau_E|
\land
F_{tran}
>
F_{res}.
\label{eq:condition_CVT}
\end{gather}
Equations~(\ref{eq:condition_AA})--(\ref{eq:condition_CVT}) indicate
that the actuation state of the STAND-MCP is governed by the force
balance between the wire-generated generalized forces and the passive
restoring elements embedded within the mechanism. According to
(\ref{eq:condition_AA}), when the FE actuation torque generated by the
bilateral wire tensions is smaller than the extension restoring torque,
FE motion remains constrained, while differential bilateral wire
actuation can generate AA motion. Once the FE actuation torque exceeds
the extension restoring torque, as defined by
(\ref{eq:condition_FE}), flexion is actuated while the passive slider
remains at its current configuration. Furthermore, under the condition
defined by (\ref{eq:condition_CVT}), the wire-generated transmission
force exceeds the slider restoring force, causing passive slider
reconfiguration. As a result, the flexion moment arm increases and the
transmission ratio is passively adapted. Consequently, the STAND-MCP
mechanically selects among AA motion, direct-transmission FE motion,
and passive CVT reconfiguration through these embedded force-balance
conditions.

\begin{figure}[t]
    \centering
    \includegraphics[width=0.5\textwidth]{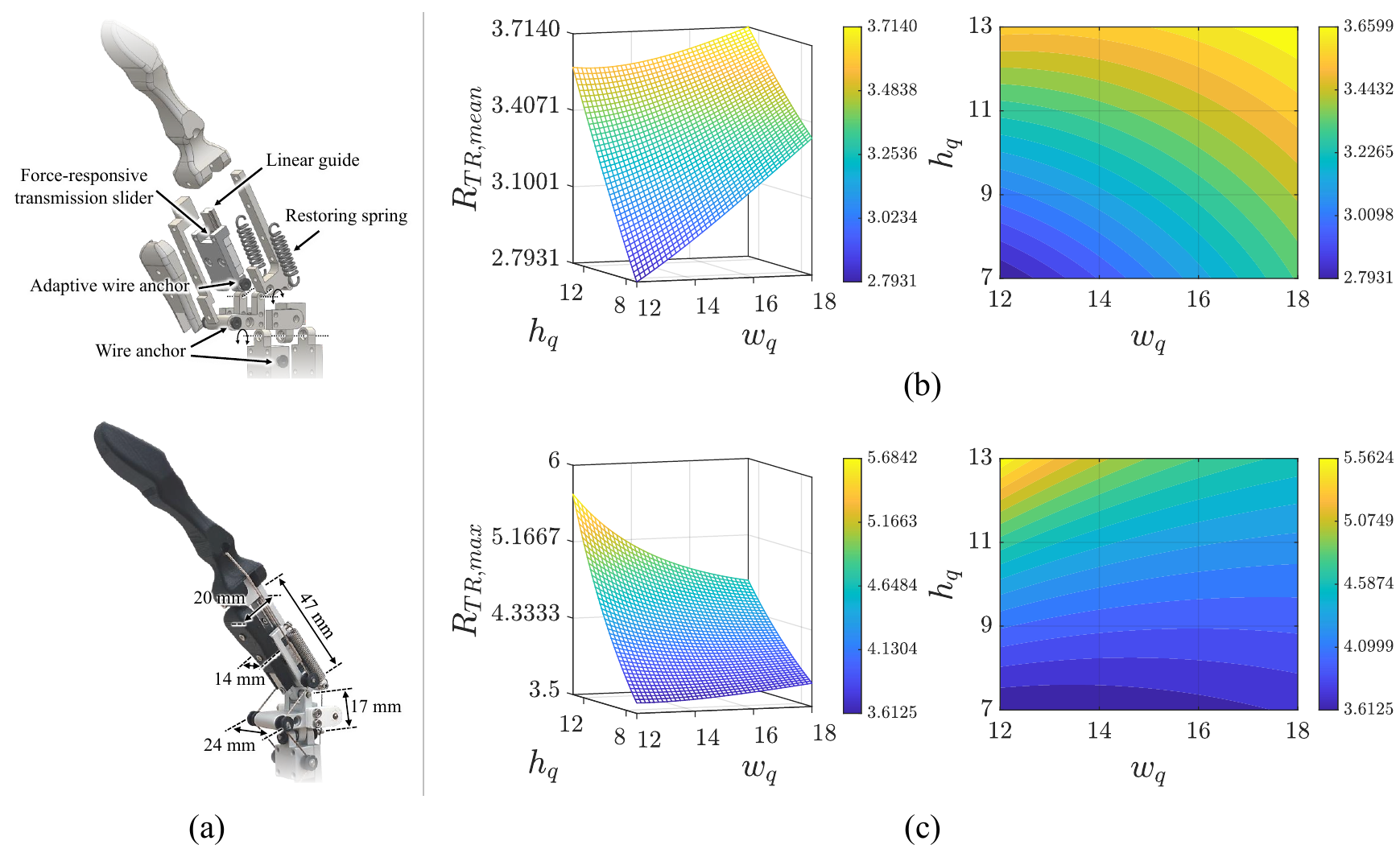}
    \caption{Overall design and design parameter analysis of the STAND-MCP. (a) Overall design and prototype of the STAND-MCP. (b) Variation of the mean transmission ratio range $R_{TR,\mathrm{mean}}$ with respect to the geometric parameters $h_q$ and $w_q$. (c) Variation of the maximum transmission ratio range $R_{TR,\max}$ with respect to the geometric parameters $h_q$ and $w_q$.}
    \label{fig:Analysis_and_comparison}
\end{figure}

\section{Design of the STAND-MCP}
\label{sec:section4}

\subsection{Overall Design and Components}
Fig.~\ref{fig:concept}(h) illustrates the STAND-MCP mechanism, which comprises four principal components: a force-responsive transmission slider, a linear guide, a restoring spring, and an adaptive wire anchor. These components collectively realize the passive CVT mechanism by geometrically reconfiguring the moment arm according to wire tension.

The force-responsive transmission slider functions as the core element of the passive CVT. During flexion actuation, increasing wire tension passively translates the slider and continuously adjusts the position of the adaptive wire anchor, thereby modulating the transmission ratio without reliance on actuators, sensors or active control. The linear guide constrains the slider to a single translational axis, ensuring stable and repeatable transmission behavior. The restoring spring governs the displacement--tension relationship of the slider and therefore determines the force condition at which transmission ratio transition occurs. The core structure excluding the wire anchor has dimensions of $47\ \mathrm{mm} \times 14\ \mathrm{mm} \times 20\ \mathrm{mm}$, whereas the overall mechanism including the wire anchor has dimensions of $47\ \mathrm{mm} \times 24\ \mathrm{mm} \times 20\ \mathrm{mm}$. The finger has a total mass of 56 g, excluding the actuators. The total finger length including the abduction/adduction joint module is 64 mm, demonstrating compactness suitable for integration into multi-DoF robotic hands.

\begin{figure*}[t]
    \centering
    \includegraphics[width=1\textwidth]{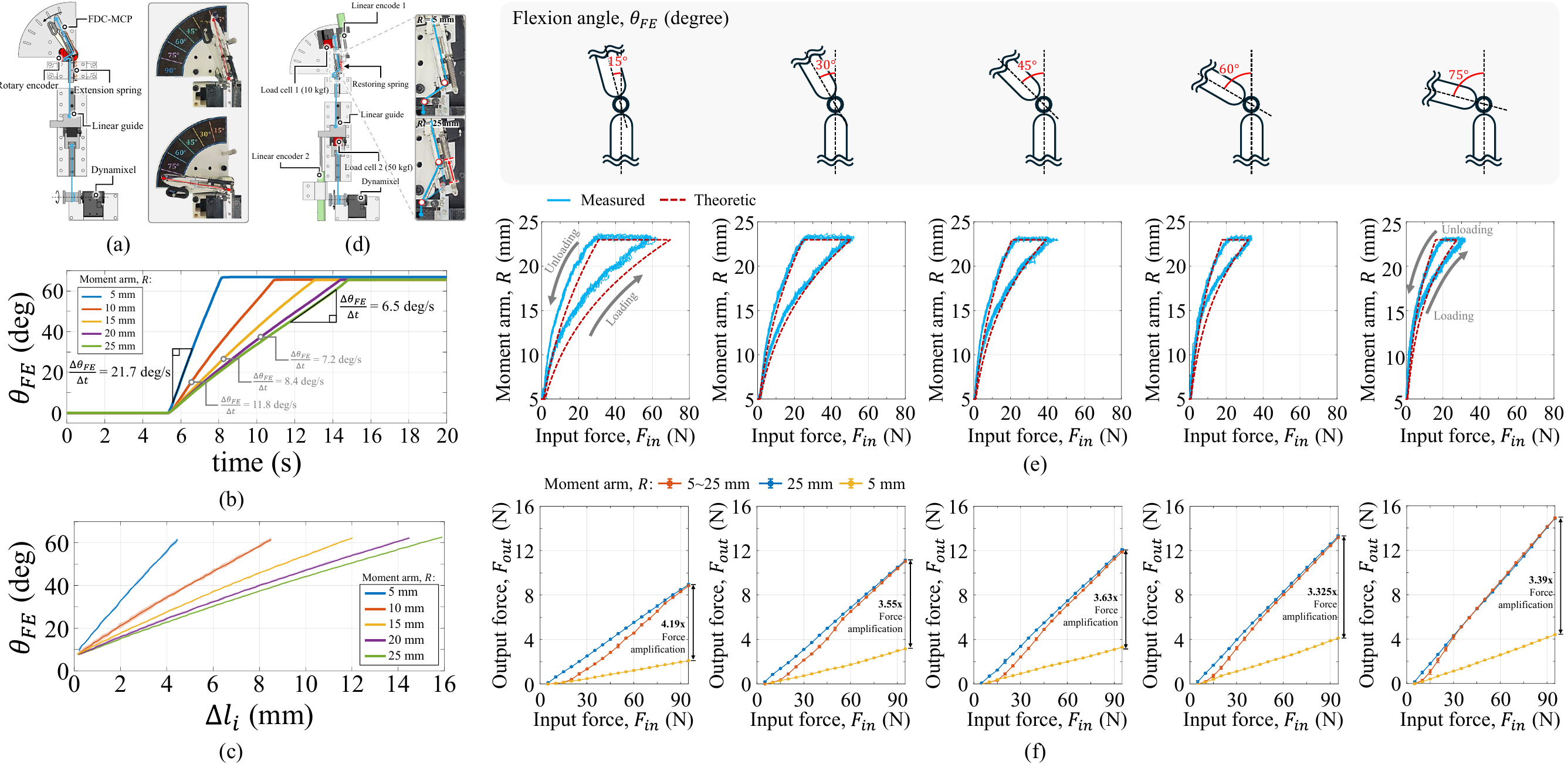}
    \caption{Experimental validation of the transmission behavior. (a) Experimental setup for flexion transmission kinematics. (b) Flexion motion speed under different moment-arm conditions. (c) Repeatability of flexion motion under repeated cycles. (d) Experimental setup for force-responsive transmission adaptation. (e) Moment-arm variation with input force at different $\theta_{FE}$ conditions during loading and unloading, compared with theoretical predictions. (f) Output-force comparison among the fixed $R=5$ mm, fixed $R=25$ mm, and passive CVT configurations at different $\theta_{FE}$ conditions.}
    \label{fig:Experiments}
\end{figure*}

\subsection{Geometric Parameter Selection}

The geometric parameters $w_q$ and $h_q$ directly govern the transmission behavior of the STAND-MCP mechanism by altering the achievable transmission ratio over the flexion moment arm $R$. To quantitatively evaluate their effects and determine suitable geometric configurations, the transmission ratio range was defined from the transmission ratio in \eqref{eq:TR_FE_i}.
For a given flexion angle $\theta_{FE}$, the transmission ratio range $R_{TR}$ is expressed as:

\begin{equation}
R_{TR}(\theta_{FE})
=
\frac{
\max_{R}|TR_{FE}(\theta_{FE},R)|
}{
\min_{R}|TR_{FE}(\theta_{FE},R)|
}
\label{eq:R_TR}
\end{equation}
This metric represents the achievable transmission modulation induced by flexion moment arm $R$ at a fixed flexion angle $\theta_{FE}$.


To evaluate the transmission characteristics throughout the flexion
workspace, $R_{TR}(\theta_{FE})$ was evaluated over the flexion moment-arm range $R\in[5,25]$ mm, with the initial moment arm set to
$R_0=5$ mm. The maximum and mean values, denoted as $R_{TR,\max}$ and $R_{TR,\mathrm{mean}}$, respectively, were then computed over $\theta_{FE}\in[5^\circ,85^\circ]$. Contour maps were generated over the geometrically feasible parameter ranges of $w_q\in[12,18]$ mm and $h_q\in[8,15]$ mm.

The resulting contour maps reveal distinct roles for each parameter (Fig.~\ref{fig:Analysis_and_comparison}(a) and (b)). Increasing $h_q$ monotonically increases both $R_{TR,\max}$ and $R_{TR,\mathrm{mean}}$, indicating that larger $h_q$ enables a wider transmission ratio range throughout the flexion workspace. However, larger $h_q$ simultaneously increases the mechanism envelope. Considering the trade-off between transmission ratio range and compactness, $h_q$ was selected as 10 mm.

In contrast, $w_q$ produces a trade-off between max and mean transmission ratio range characteristics. Increasing $w_q$ improves $R_{TR,\mathrm{mean}}$ while reducing $R_{TR,\max}$, implying that the geometric design determines whether broader mean transmission ratio modulation or max transmission ratio amplification is prioritized. Considering the spatial constraints of the finger mechanism, $w_q$ was selected as 15 mm.

\subsection{Extension Spring Stiffness Selection}
The extension spring parameters were determined to satisfy the restoring torque required for stable finger extension while considering the inertial and gravitational loading of the finger mechanism. To conservatively estimate the required torque, the finger link was modeled as a rigid aluminum block with dimensions of $20\ \mathrm{mm} \times 20\ \mathrm{mm} \times 40\ \mathrm{mm}$, yielding a mass of $m_f \approx 43.2\ \mathrm{g}$ and a rotational inertia of $I_f \approx 2.45 \times 10^{-5}\ \mathrm{kg \cdot m^2}$ about the flexion--extension axis.

The gravitational torque was evaluated at the worst-case configuration corresponding to the fully extended horizontal posture, resulting in $\tau_g \approx 8.48 \ \mathrm{N \cdot mm}$. Assuming a maximum angular velocity of $\omega_{\max}=200\ \mathrm{deg/s}$ and uniform acceleration over the full flexion range of $85^\circ$, the corresponding angular acceleration and inertial torque were estimated as $\alpha_{\max}\approx4.10\ \mathrm{rad/s^2}$ and $\tau_I\approx1.00\times10^{-1}\ \mathrm{N \cdot mm}$, respectively. Accordingly, the minimum required extension torque was estimated as 
$\tau_{E,re}=\tau_g+\tau_I \approx 8.58\ \mathrm{N \cdot mm}$.

Based on this criterion, the spring parameters were selected to ensure that the extension torque remained greater than $\tau_{E,re}$ throughout the entire $\theta_{FE}$ range. Considering commercially available extension springs, the spring stiffness and pre-tension force were determined as $k_E=0.11\ \mathrm{N/mm}$ and $F_{E,\mathrm{pre}}=1\ \mathrm{N}$, respectively. The remaining geometric parameters were selected as $a_E=10\ \mathrm{mm}$, $b_E=3\ \mathrm{mm}$, $w_E=8\ \mathrm{mm}$, and $h_E=8\ \mathrm{mm}$.

\subsection{Restoring Spring Stiffness Selection}

The restoring spring parameters were determined to ensure stable flexion behavior 
while preventing unintended variation of the flexion moment arm $R$ during 
flexion motion.

The restoring spring was required to generate sufficient restoring force to overcome the extension torque throughout the operating range such that the flexion input force would not induce undesired displacement of $R$. This condition necessitates a sufficiently large initial restoring force $F_{res,0}$. As calculated from \eqref{eq:extension_torque}, the maximum extension torque was approximately 16 N·mm and the spring parameters were selected accordingly. In addition, the restoring spring was required to accommodate a deflection of more than 20 mm while remaining functional throughout the operating range.

Considering these requirements, $F_{res,0}$ is sufficiently large to satisfy the initial restoring force requirement, the selection of $k_{res}$ determines the restoring force profile over the operating range and can therefore be adjusted according to the design objectives. Accordingly, the restoring spring parameters were determined as $k_{res}=0.6\ \mathrm{N/mm}$ ($2 \times 0.3 \ \mathrm{N/mm}$) and $F_{res,0}=0.9\ \mathrm{N}$.

\section{Experimental Validation}
\label{sec:section5}

\subsection{Experimental Validation of Flexion Transmission Kinematics}
Flexion experiments were conducted to evaluate the effect of the moment arm radius on the flexion transmission ratio of the STAND-MCP mechanism. The setup consists of a rotary encoder (AMT103, CUI Devices), a linear guide, and a Dynamixel servo motor (XM540-W270-R, Robotis), as shown in Fig.~\ref{fig:Experiments}(a). The motor drove the finger at a constant input speed of 60 deg/s with a 7 mm pulley radius, while the linear guide constrained the transmission slider to pure translational motion. Five cases of fixed moment arm radius ($R$ = 5, 10, 15, 20, and 25 mm) were tested independently under identical actuator input conditions. All sensor signals were sampled at 30 Hz throughout the experiments.

Fig.~\ref{fig:Experiments}(b) shows the time required for $\theta_{FE}$ to reach 65$^\circ$ from 0$^\circ$ under each $R$ condition, illustrating the effect of moment arm radius on output motion speed. The required time increased with $R$, and the reduction in output angular velocity was more significant in the smaller $R$ region, indicating a nonlinear tendency. Quantitatively, the average flexion angular velocity decreased from 21.7 deg/s at $R$ = 5 mm to 6.5 deg/s at $R$ = 25 mm, representing an approximately 3.3-fold decrease. This corresponds to a proportional increase in the effective transmission ratio under identical actuator input. These results experimentally verify, at the kinematic level, that the proposed passive CVT mechanism can continuously vary the transmission ratio by changing the moment arm geometry.
To examine repeatability, an additional experiment was conducted in which $\theta_{FE}$ cycled between 5$^\circ$ and 65$^\circ$ for five repetitions under the same experimental conditions. Fig.~\ref{fig:Experiments}(c) presents the flexion angle as a function of wire displacement, where the solid lines indicate the mean values over the five repetitions and the shaded regions represent the corresponding standard deviations. The standard deviations remain small over the entire motion range and closely follow the mean trajectories, indicating highly consistent kinematic responses across repeated cycles. These results demonstrate that the STAND-MCP provides repeatable transmission performance under the tested conditions.

\subsection{Experimental Validation of Force-Responsive Transmission Behavior}

\begin{figure*}[t]
    \centering
    \includegraphics[width=0.95\textwidth]{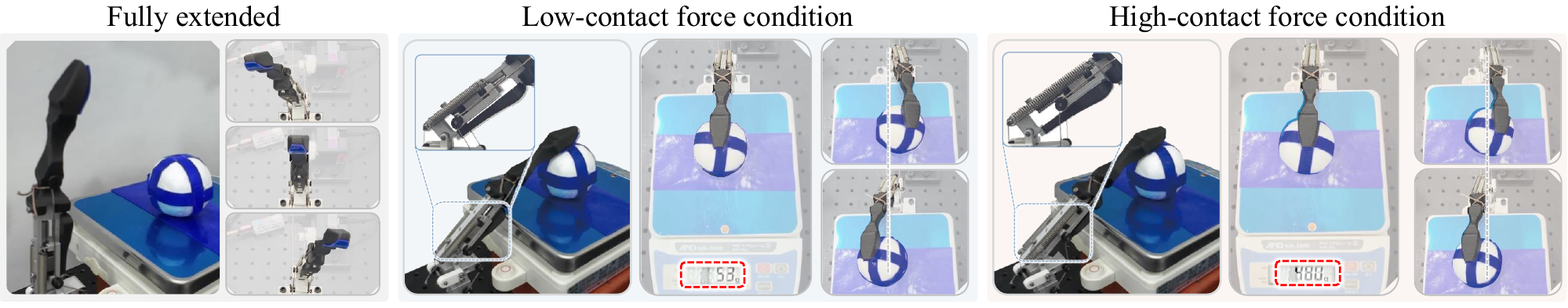}
    \caption{Dexterous ball-rolling demonstrations under three conditions: non-contact abduction/adduction, low-contact-force rolling (flexed posture, ~0.52 N (53 g)), and high-contact-force rolling (flexed posture, ~4.71 N (480 g)).}
    \label{fig:Demo}
\end{figure*}
The proposed passive CVT mechanism is designed to adaptively vary the moment arm $R$ and increase the output force as the input force increases. To validate this force-responsive transmission behavior, the moment arm variation and output force amplification were experimentally evaluated under different input force and joint angle conditions. All sensor signals were sampled at 30 Hz throughout the experiments.

Fig.~\ref{fig:Experiments}(d) illustrates the experimental setup, consisting of a Dynamixel servo motor (XM540-W270-R, Robotis), two linear encoders (PZ34-A-150, GEFRAN), load cell 1 (UU3, Dacell, 10 kgf), and load cell 2 (UU3, Dacell, 50 kgf). Load cells 1 and 2 have a nonlinearity and hysteresis of 0.05\% of rated output and a repeatability of 0.03\% of rated output. The Dynamixel applied input force to the wire. Linear encoder 1 measured the effective flexion moment arm $R$ through the displacement of the force-responsive transmission slider, whereas linear encoder 2 measured the input wire displacement. Load cell 1 measured the output force at the fingertip, and load cell 2 measured the input wire force. Experiments were conducted under different fixed $\theta_{FE}$ conditions to evaluate the variation of $R$ and the corresponding output force amplification with increasing input force.


Fig.~\ref{fig:Experiments}(e) shows the variation of the moment arm radius $R$ as a function of input force under different $\theta_{FE}$ conditions. For each $\theta_{FE}$ condition, the experiment was repeated 10 times. For all $\theta_{FE}$ conditions, $R$ increased with increasing input force, and its sensitivity to the input force increased as $\theta_{FE}$ increased. The theoretical curves were fitted using effective friction coefficients of $\mu^{+}=0.01$ for loading and $\mu^{-}=0.3$ for unloading. With these coefficients, the theoretical predictions showed close agreement with the experimental results for both loading and unloading, except for the $\theta_{FE}=15^\circ$ condition. These results indicate that transmission reconfiguration depends on finger posture and confirm that the proposed mechanism passively reconfigures its transmission geometry in response to the applied input force. The overall agreement between the theoretical and experimental results further supports the validity of the proposed Jacobian-based transmission model.

Fig.~\ref{fig:Experiments}(f) compares the measured output forces of two fixed-radius systems ($R=5$ mm and $R=25$ mm) and the proposed passive CVT system under different $\theta_{FE}$ conditions. During the experiment, the input force was increased by 5 N every 10 s while the corresponding output force was measured.
The output-force amplification ratios of the fixed $R=25$ mm configuration relative to the fixed $R=5$ mm configuration were approximately 4.19, 3.55, 3.65, 3.35, and 3.39 at $\theta_{FE}=15^\circ,\ 30^\circ,\ 45^\circ,\ 60^\circ,$ and $75^\circ$, respectively. Accordingly, the maximum and mean output-force amplification ratios across the five tested flexion angles from $\theta_{FE}=15^\circ$ to $\theta_{FE}=75^\circ$ were approximately 4.19 and 3.63, respectively.

These results indicate that the force-transmission characteristics of the system vary substantially depending on the flexion moment-arm radius.
The passive CVT initially exhibited output-force characteristics similar to those of the $R=5$ mm configuration. As the input force increased, the flexion moment arm gradually expanded, resulting in a continuous increase in transmission ratio and output force. As a result, the output-force characteristics progressively approached those of the $R=25$ mm configuration.




\section{Demonstration}
\label{sec:section6}
To evaluate dexterous manipulation capability, ball-rolling demonstrations were conducted across three scenarios shown in Fig.~\ref{fig:Demo}: (1) abduction/adduction motion in the fully extended configuration without object contact as a kinematic baseline, (2) rolling manipulation under a low-contact-force condition (flexed posture, approximately 0.52 N (53 g)), and (3) rolling manipulation under a high-contact-force condition (flexed posture, approximately 4.71 N (480 g)), where the passive CVT engaged to increase the effective transmission ratio. Experimental results confirmed stable abduction/adduction motion in all conditions. Successful ball rolling manipulation was achieved under both contact conditions despite the substantially different contact force levels. These results demonstrate that the proposed mechanism passively adapts its transmission characteristics according to contact conditions while preserving dexterous lateral manipulation capability.

\section{DISCUSSION AND CONCLUSIONS}
\label{sec:section7}
This study presented the STAND-MCP, a compact 2-DoF robotic MCP mechanism that realizes DoF-specific transmission differentiation to address the force–speed tradeoff in compact robotic fingers. By introducing an output-side passive CVT architecture, the proposed mechanism enables transmission adaptation to be selectively assigned to the flexion DoF while preserving the compact and direct transmission characteristics required for abduction–adduction. This approach allows transmission adaptation to be incorporated within a multi-DoF MCP architecture without additional actuators, sensors, or control.

Experimental results demonstrated repeatable passive transmission reconfiguration, with a maximum output-force amplification of 4.19-fold and a mean amplification of 3.63-fold across five tested flexion angles from $\theta_{FE}=15^\circ$ to $\theta_{FE}=75^\circ$. Stable force-responsive transmission behavior was verified across varying input-force and joint-angle conditions. Furthermore, ball-rolling demonstrations confirmed that coordinated 2-DoF MCP motion was preserved during passive transmission ratio reconfiguration of the flexion DoF.

This work demonstrates that transmission ratio behavior can be selectively reconfigured at the transmission-geometry level, providing a new design direction for compact and scalable robotic hands. However, several limitations remain. Human-level force–speed performance has not yet been evaluated with compact actuators fully integrated within the hand. In addition, AA performance, extension to higher-DoF finger architectures, and the resulting workspace have not been systematically evaluated. The dynamic behavior of the system requires further investigation, as the spring-dependent extension mechanism can affect control performance under high-bandwidth operation. Finally, long-term lifecycle, wear, and robustness have not yet been assessed. Future work will address these limitations through fully integrated in-hand actuation, higher-DoF workspace and dynamic-performance evaluation, and long-term durability testing.

\addtolength{\textheight}{-12cm}   







\bibliographystyle{IEEEtran}
\bibliography{references}

\end{document}